\documentclass[11pt, a4paper, logo, copyright, nonumbering]{map}

\usepackage{CJKutf8}
\usepackage{xargs}  
\usepackage{todonotes}
\usepackage{amsmath}
\usepackage{dsfont}

\usepackage{longtable}

\usepackage{svg}
\usepackage{fontawesome}
\usepackage{array}
\usepackage{tabularx}
\usepackage{latexsym}
\usepackage{graphicx}
\usepackage[utf8]{inputenc} 
\usepackage[utf8]{inputenc} 
\usepackage{amssymb} 
\usepackage{url}            
\usepackage{amsfonts}       
\usepackage{nicefrac}       
\usepackage{microtype}      
\usepackage{xspace}

\usepackage{listings}
\usepackage{makecell}
\usepackage{inconsolata}
\usepackage{multicol}
\usepackage{amstext}

\definecolor{darkblue}{RGB}{84, 112, 198}

\definecolor{lightblue}{rgb}{0.85, 0.95, 1.0}    
\definecolor{lightgreen}{rgb}{0.90, 1.0, 0.90}    
\definecolor{lightorange}{rgb}{1.0, 0.95, 0.85}   
\definecolor{lightpurple}{rgb}{0.95, 0.90, 1.0}   
\definecolor{lightgray}{rgb}{0.97, 0.97, 0.97}    

\usepackage{tikz}
\usetikzlibrary{calc}
\definecolor{battery-empty}{rgb}{0.9, 0.9, 0.9}
\newcommand{\difficultybar}[1]{%
  \begin{tikzpicture}[baseline, scale=0.5, every node/.style={scale=0.8}]
    \foreach \i in {1,2,3,4,5} {
      \ifnum\i>#1
        \draw[fill=battery-empty] (\i*0.5-0.5, 0) rectangle (\i*0.5, 0.25);
      \else
        \pgfmathsetmacro{\colorlevel}{80 - 12*(\i)} 
        \edef\x{\noexpand\draw[fill=blue!\colorlevel!white, opacity=0.9] (\i*0.5-0.5, 0) rectangle (\i*0.5, 0.25);}
        \x
        \draw[blue!50!black] (\i*0.5-0.5, 0) rectangle (\i*0.5, 0.25);
      \fi
    }
    \fill[battery-empty!70] (2.5, 0.08) rectangle (2.6, 0.17);
    \draw[battery-empty!70!black] (2.5, 0.08) rectangle (2.6, 0.17);
  \end{tikzpicture}%
}

\definecolor{hidden-draw}{RGB}{20,68,106}
\definecolor{hidden-pink}{RGB}{255,245,247}
\usepackage[edges]{forest}

\usepackage{natbib}
\usepackage{CJKutf8}
\usepackage{xargs}
\usepackage{todonotes}
\usepackage{multirow}
\usepackage{cleveref}
\usepackage{dsfont}
\usepackage{subcaption}
\usepackage{url}
\usepackage{svg}
\usepackage{fontawesome}
\usepackage{xcolor}
\usepackage{float}

\usepackage[utf8]{inputenc}
\usepackage{booktabs}
\usepackage{graphicx}
\usepackage{array}
\usepackage{tabularx}
\usepackage{xcolor,colortbl}  
\definecolor{boxcolor}{HTML}{d92523} 
\definecolor{bulbcolor}{HTML}{e3b87f} 
\usepackage{url}
\usepackage{ragged2e}   
\usepackage{makecell} 

\usepackage{hyperref}       
\usepackage{url}            
\usepackage{booktabs}       
\usepackage{amsfonts}       
\usepackage{nicefrac}       
\usepackage{microtype}      
\usepackage{xcolor}         
\usepackage{graphicx}
\usepackage{longtable}
\usepackage{array}
\usepackage{pbox}
\usepackage{tabularx}
\usepackage{multirow}
\usepackage{wrapfig}
\usepackage{pifont}
\usepackage{algorithm}
\usepackage{lipsum}
\usepackage{subcaption}
\usepackage{enumitem}
\usepackage{tikz}
\usepackage{xstring}
\usepackage{bm}
\usepackage{color-edits}
\usepackage{float}
\usepackage{parskip} 

\newcommand{\hidden}{\mathbf{h}}
\newcommand{\block}{f_\theta}
\newcommand{\loopcount}{R}
\newcommand{\nparams}{N}

\usepackage{threeparttable} 

\newcommand{\modelname}{LoopCoder-v2} 
\usepackage{algorithm}
\usepackage{algpseudocode}

\algnewcommand\algorithmicinput{\textbf{Input:}}
\algnewcommand\algorithmicoutput{\textbf{Output:}}
\algnewcommand\Input{\item[\algorithmicinput]}
\algnewcommand\Output{\item[\algorithmicoutput]}

\definecolor{rliableolive}{HTML}{BBCC33}
\definecolor{rliableblue}{HTML}{77AADD}
\definecolor{rliablered}{HTML}{f63c44}

\usepackage[most,skins,theorems]{tcolorbox}
\tcbuselibrary{skins,breakable,fitting,hooks}  
\tcbset{
  aibox/.style={
    width=\linewidth,
    top=7pt,
    bottom=2pt,
    colback=rliablered!18!white,
    colframe=black,
    colbacktitle=black,
    enhanced,
    center,
    attach boxed title to top left={yshift=-0.1in,xshift=0.15in},
    boxed title style={boxrule=0pt,colframe=white,},
  }
}
\newtcolorbox{AIbox}[2][]{aibox,title=#2,#1}

\addauthor[Zhejian]{zzj}{blue}

\hypersetup{
    colorlinks=true,            
    linkcolor=blue,             
    filecolor=magenta,          
    urlcolor=cyan,              
    citecolor=purple,             
    pdftitle={Overleaf Example},
    pdfpagemode=FullScreen,
}

\renewcommand{\today}{}
\newcommandx{\info}[2][1=]{\todo[linecolor=red,backgroundcolor=red!25,bordercolor=red,#1]{#2}}

\title{
\centering{\modelname{}: Only Loop Once for Efficient Test-Time Computation Scaling}
}

\author{
\centering
\textbf{Jian Yang}\textsuperscript{1} 
\textbf{Shawn Guo}\textsuperscript{2}
\textbf{Wei Zhang}\textsuperscript{2}
\textbf{Tianyu Zheng}\textsuperscript{2}
\textbf{Yaxin Du}\textsuperscript{2}
\textbf{Haau-Sing Li}\textsuperscript{2}
\textbf{Jiajun Wu}\textsuperscript{2} 
\textbf{Yue Song}\textsuperscript{2}
\textbf{Yan Xing}\textsuperscript{2}
\textbf{Qingsong Cai}\textsuperscript{2}
\textbf{Zelong Huang}\textsuperscript{2}
\textbf{Chuan Hao}\textsuperscript{2}
\textbf{Ran Tao}\textsuperscript{2}
\textbf{Xianglong Liu}\textsuperscript{1}
\textbf{Wayne Xin Zhao}\textsuperscript{4}
\textbf{Mingjie Tang}\textsuperscript{2}
\textbf{Weifeng Lv}\textsuperscript{1}
\textbf{Ming Zhou}\textsuperscript{3} 
\textbf{Bryan Dai}\textsuperscript{2} \\
\textsuperscript{1}Beihang University
\textsuperscript{2}IQuest Research
\textsuperscript{3}Langboat
\textsuperscript{4}Renming University of China \\
\texttt{\{jiayang\}@buaa.edu.cn}

{
\includegraphics[height=1em]{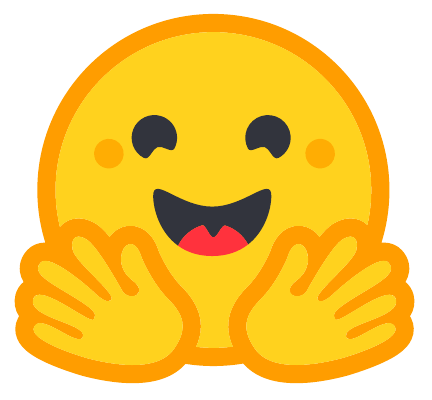}\;
HuggingFace: \url{https://huggingface.co/Multilingual-Multimodal-NLP/LoopCoder-V2}
}\\
\vspace{-20pt}
}

\begin{abstract}
Looped Transformers scale latent computation by repeatedly applying shared blocks, but sequential looping increases latency and KV-cache memory with the loop count. Parallel loop Transformers (PLT) alleviate this cost through cross-loop position offsets (CLP) and shared-KV gated sliding-window attention, making loop count a practical design choice. We therefore study PLT loop-count selection through a gain--cost view: an extra loop may refine representations, but CLP also introduces a positional mismatch at each loop boundary. We instantiate this study by training LoopCoder-v2, a family of 7B PLT coders with different loop counts, from scratch on 18T tokens, followed by matched instruction tuning and evaluation. Empirically, the two-loop variant delivers broad gains over the non-looped baseline across code generation, code reasoning, agentic software engineering, and tool-use benchmarks, improving SWE-bench Verified from 43.0 to 64.4 points and Multi-SWE from 14.0 to 31.0 points. In contrast, variants with three or more loops regress, revealing a strongly non-monotonic loop-count effect. Our diagnostics show that loop 2 provides the main productive refinement, while later loops yield diminishing, oscillatory updates and reduced representational diversity. Because the CLP-induced mismatch remains roughly fixed as refinement gains shrink, the offset cost increasingly dominates. This gain--cost trade-off explains PLT's saturation at two loops and provides diagnostics for loop-count selection.
\end{abstract}

\begin{document}
\maketitle

\section{Introduction}
Looped large language models (LLM)~\cite{yang2025codesurvey,wu2025parallel} have emerged as a promising way to scale the effective computational depth of language models without proportionally increasing their parameter count. Instead of stacking many distinct layers, a looped large language model (LLM) repeatedly applies a shared Transformer block, allowing the same parameters to perform multiple rounds of latent-space computation \citep{dehghani2018universal,giannou2023looped,yang2023looped}. This design is especially attractive for test-time compute scaling, enabling additional internal refinement without generating auxiliary reasoning tokens \citep{geiping2025scaling}. Recent work has shown that such recurrent-depth LLMs can approach deeper non-looped Transformers and improve reasoning performance as more inference-time computation is used \citep{geiping2025scaling,yang2026stabilizing,schwethelm2026how}.

Despite this promise, standard sequential looping is difficult to deploy efficiently: each additional loop requires another pass through the shared block and introduces loop-specific KV-cache states, causing both latency and memory to grow with the loop count \citep{vendrell2026memoryefficient}. The Parallel Loop Transformer (PLT) \citep{wu2025parallel} mitigates this bottleneck with two complementary mechanisms: cross-loop position offsets (CLP), which break sequential inter-loop dependencies and enable parallel loop execution, and shared-KV gated sliding-window attention (G-SWA), which keeps the cache footprint nearly constant across loop counts. Yet reducing the cost of looping does not by itself determine the best operating point. In PLT, the loop count becomes a deployment-time design choice: too few loops may underuse the model's refinement capacity, while too many loops may introduce redundant or harmful computation. Exhaustively training and evaluating every candidate loop count is expensive and offers little insight into why a particular setting succeeds or fails. This motivates a more diagnostic question: \emph{can we identify the saturation point of PLT by examining what each loop contributes internally?}

\begin{figure}[t]
\centering
\includegraphics[width=0.99\linewidth]{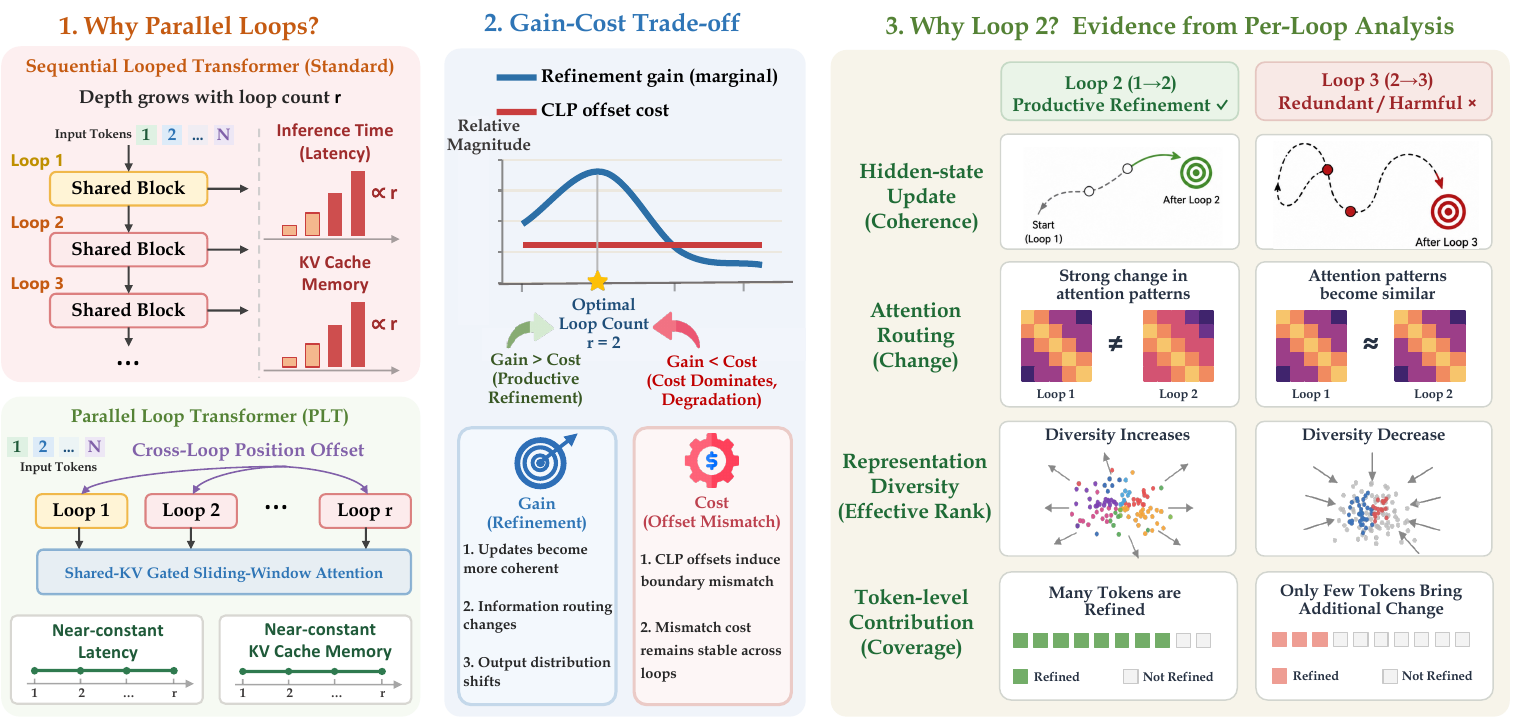}
\caption{Overview of PLT loop-count selection. Left: standard sequential looping
increases latency and KV-cache memory with the loop count, whereas PLT uses a
cross-loop position offset and shared-KV G-SWA to keep both costs nearly
constant. Middle: each added loop trades marginal refinement gain against the
CLP-induced offset mismatch. The gain peaks early and then shrinks while the
offset cost remains roughly stable, making $r=2$ the preferred operating point.
Right: per-loop diagnostics explain this choice. Loop~2 shows coherent
hidden-state movement, changed attention routing, increased representation
diversity, and broad token-level refinement, while loop~3 is more redundant and
less productive.}
\label{fig:main}
\end{figure}

To investigate this question, we view PLT's loop-count behavior through a gain--cost lens (\autoref{fig:main}). On the gain side, an additional loop is useful only if it performs meaningful refinement: coherently updating hidden states, changing information routing, and shifting the model's output distribution. We therefore track hidden-state dynamics, attention evolution, and output-distribution shift across loops. On the cost side, CLP enables parallelism by replacing direct same-token recurrence with an offset dependence on neighboring states, which can introduce a positional mismatch at each loop boundary. We quantify this mismatch from the model's hidden states and relate it to the marginal gain of each loop. All comparisons are conducted under matched training, instruction-tuning, and evaluation settings, ensuring that the resulting loop-wise differences reflect the effect of loop count rather than changes in protocol.

We instantiate this study on \modelname{}, a 7B PLT coder trained from scratch on 18T tokens of mixed text and code data, followed by instruction tuning. Comparing matched loop-count variants with (\loopcount $\in \{1,2,3,4\}$), we observe a strongly non-monotonic trend: the two-loop model improves broadly over the non-looped baseline, including a gain on SWE-bench Verified from 43.0\% to 64.4\%, while the three-loop model regresses on many tasks, dropping to 27.6\% on SWE-bench Verified. This pattern indicates that additional PLT loops can become harmful, motivating our loop-wise analysis of marginal refinement gain and CLP-induced offset cost.

Our contributions are as follows:

\begin{enumerate}
\item \textbf{A gain--cost view of PLT loop-count selection.}
We formulate PLT loop-count selection as a trade-off between the marginal refinement gained from additional loops and the structural cost introduced by CLP at loop boundaries.

\item \textbf{A loop-wise diagnosis of PLT saturation.}
We analyze hidden-state dynamics, attention evolution, and output-distribution shift across loops, and define an intrinsic offset cost $\Omega^{(r)}$ to quantify CLP-induced positional mismatch. Our analysis shows that the second loop provides the main productive refinement, while later loops yield diminishing and increasingly oscillatory updates.

\item \textbf{A large-scale empirical study with a 7B PLT coder.}
We train \modelname{} from scratch with different loop counts on 18T tokens of mixed text and code data and compare loop-count variants under matched training, instruction-tuning, and evaluation settings. The two-loop model improves broadly across code generation, code reasoning, agentic software engineering, and tool-use benchmarks, while additional looping often regresses.
\end{enumerate}

\section{Preliminaries and Problem Formulation}
\label{sec:plt}
This section formalizes the loop-count selection problem studied in this paper. We first review standard looped Transformers, where additional loops increase latent computational depth but also incur sequential inference cost. We then introduce the parallel loop Transformer (PLT), which reduces this cost through parallel loop execution and shared-KV attention, while changing the information flow across loop iterations. Finally, we compare the two settings and formulate PLT loop-count selection as a gain--cost trade-off: each additional loop may provide useful representational refinement, but in PLT it also introduces an offset-induced positional mismatch.

\subsection{Looped Transformers}
\label{sec:looped}
A looped Transformer replaces a deep stack of distinct layers with a single
shared block $\block$ of $L$ layers applied repeatedly
\citep{dehghani2018universal,giannou2023looped}.
Given input tokens $x$, it unrolls the recurrence
\begin{equation*}
  \hidden^{(0)} = \mathrm{Embed}(x),
  \qquad
  \hidden^{(r)} = \block\!\left(\hidden^{(r-1)}\right),\;\; r = 1,\dots,\loopcount,
  \qquad
  logits = \mathrm{Head}\!\left(\hidden^{(\loopcount)}\right),
  \label{eq:loop-rec}
\end{equation*}
where $\hidden^{(r)}$ is the $r$-th hidden state, $\loopcount$ the loop counts.
Reusing the same parameters at every iteration makes the effective depth
$\loopcount\cdot L$ grow with $\loopcount$ while the parameter count $\nparams$
stays fixed, so a looped model attains deep computation on a small parameter
budget, rivaling much larger non-looped models on depth-sensitive tasks
\citep{yang2023looped}.

This depth, however, comes at an inference cost that scales with $\loopcount$
(\autoref{tab:loop-vs-plt}).
The recurrence is strictly \emph{sequential}: $\hidden^{(r)}$ cannot be computed
before $\hidden^{(r-1)}$, so $\loopcount$ loops require $\loopcount$ successive
passes through $\block$ and multiply wall-clock latency by $\loopcount$.
The memory cost grows just as steeply: a standard implementation caches the keys
and values of every layer \emph{at every loop}, so the KV-cache footprint reaches
$O(\loopcount \cdot L \cdot S \cdot d)$ for sequence length $S$ and width $d$, a
factor of $\loopcount$ over a single pass \citep{vendrell2026memoryefficient}.
Both latency and memory thus grow with every added loop, making deeply looped
inference impractical in latency- or memory-constrained deployment.

\subsection{Parallel Loop Transformer}
\label{sec:plt-arch}
The parallel loop Transformer (PLT) \citep{wu2025parallel} is a representative design that reduces these two costs. PLT introduces two independent mechanisms: a shared first-loop KV cache accessed through gated sliding-window attention, which bounds memory, and a cross-loop position offset,
which removes the sequential dependency between loops.

\paragraph{Efficient Representation Enhancement (KV sharing + G-SWA).}
The KV cache from the first loop,
$K_{\text{share}}, V_{\text{share}} = \mathrm{KV}(\hidden^{(1)})$,
is shared with all subsequent loops to keep total KV-cache memory at
$O(L \cdot S \cdot d)$ regardless of $\loopcount$.
In non-first loops, each attention layer performs a gated fusion of two
attention outputs:
\begin{equation}
  \tilde{y}^{(r)} = g \odot y_{\text{global}}^{(r)}
                   + (1 - g) \odot y_{\text{local}}^{(r)},
  \quad g = \sigma\!\left(f_{\text{gate}}(\mathrm{RMSNorm}(\hidden))\right),
  \label{eq:gswa}
\end{equation}
where $y_{\text{global}}^{(r)}$ is full-context attention on the frozen
$K_{\text{share}}, V_{\text{share}}$ from loop 1, and $y_{\text{local}}^{(r)}$
is sliding-window attention of width $w = 64$ over the current loop's KV. The gate $f_{\text{gate}}$ is a head-wise linear layer applied to the
RMSNorm-normalized layer input, producing one scalar per head.
\paragraph{Cross-Loop Parallelism (CLP offset).}
Before each loop $r \geq 2$, the previous loop's hidden states are
right-shifted by one token position and added back to the input:
\begin{equation}
  B^{(r)} = \mathrm{Embed}(x) + \mathrm{shift}\!\left(\hidden^{(r-1)}\right),
  \qquad
  \hidden^{(r)} = \block\!\left(B^{(r)}\right),
  \label{eq:clp}
\end{equation}
where $\mathrm{Embed}(x)$ is the token-embedding sequence of input $x$,
$\hidden^{(r-1)}$ the loop-$(r-1)$ hidden states, and $\mathrm{shift}$ the
one-position right shift $\mathrm{shift}(\hidden^{(r-1)})_i=\hidden^{(r-1)}_{i-1}$
($\hidden^{(r-1)}_{0}=\mathbf{0}$). The sum $B^{(r)}$ is the input to loop $r$,
which the shared block $\block$ maps to $\hidden^{(r)}$.
This removes the direct positional dependency between states at the same
index across consecutive loops, so the $r$-th loop of token $x_i$ can be
computed concurrently with the $(r+1)$-th loop of token $x_{i-1}$ within a
single forward pass, yielding near-single-pass wall-clock latency.

\paragraph{Information-flow consequence of the offset.}
Token $x_i$ at loop $r \geq 2$ receives as input a mixture of its own
embedding and the loop-$(r-1)$ hidden state of token $x_{i-1}$ rather
than its own.
The offset therefore induces a per-token positional mismatch: the state
available to token $x_i$ at loop $r$ reflects the context seen by
$x_{i-1}$, not $x_i$ itself.
This positional mismatch is the core information constraint that PLT introduces.

\subsection{Loop-Count Selection as a Gain--Cost Trade-off}
\label{sec:gaincost}

\autoref{tab:loop-vs-plt} contrasts the two settings: PLT removes the sequential
dependency and freezes the first-loop KV cache, so neither latency nor memory
scales with $\loopcount$.
These savings make additional loops affordable at inference, and we adopt PLT as
the experimental framework for the remainder of this paper.

\begin{table}[t]
\centering
\caption{Sequential looping versus PLT, where $C_{\text{block}}$ is the cost of
one pass through the shared block. PLT keeps both latency and memory independent
of the loop count $\loopcount$.}
\label{tab:loop-vs-plt}
\begin{tabular}{lll}
\toprule
 & \textbf{Sequential loop} & \textbf{PLT} \\
\midrule
Execution        & sequential                        & parallel, single pass \\
Latency          & $O(\loopcount\,C_{\text{block}})$  & $\approx C_{\text{block}}$ \\
KV-cache         & $O(\loopcount L S d)$              & $O(L S d)$ \\
Inter-loop input & $\hidden^{(r-1)}$                  & $\mathrm{Embed}(x) + \mathrm{shift}(\hidden^{(r-1)})$ \\
\bottomrule
\end{tabular}
\end{table}

However, this efficiency is not free: the CLP offset 
makes every added loop couple a representational \emph{gain},further latent
refinement with a positional-mismatch \emph{cost} that standard looping never
incurs.
How these opposing forces balance sets the loop count at which PLT performs
best, the gain--cost trade-off that the rest of the paper quantifies through a
per-loop interpretability analysis (\autoref{sec:analysis}).

\section{Analyzing Parallel Loop Transformers}
\label{sec:experiments}

We analyze PLT's loop-count behavior at two complementary resolutions.
We first take a \emph{macroscopic} view : under a
strictly matched protocol we vary the loop count and measure its effect on
downstream performance, establishing \emph{that} the loop count matters and
\emph{where} the best operating point lies.
We then take a \emph{microscopic} view: a battery
of per-loop diagnostic lenses that open up the model's internal computation and
explain \emph{why} the macroscopic curve takes the shape it does.
The first view tells us \emph{what} happens as loops accumulate. The second
tells us \emph{what each loop is actually doing} to produce that outcome.

\subsection{Macroscopic View: The Loop-Count Effect}
\label{sec:exp-macro}

We begin by isolating the effect of the loop count itself.
Holding the architecture, data, and tuning fixed and varying only
$\loopcount$, we ask how downstream task performance responds as we spend more
latent loops at inference, the macroscopic phenomenon that the microscopic
analysis in \autoref{sec:exp-micro} is designed to explain.

\paragraph{Model Configuration.}
\label{sec:exp-arch}

All analyses are conducted on a 7B-parameter dense transformer
equipped with the PLT mechanism, G-SWA with window size $w = 64$ and
first-loop KV sharing, applied uniformly across all loops.
Full architecture configurations are documented in
\autoref{app:model-config}.

\paragraph{Training Protocol.}
\label{sec:exp-train}

Models are trained on an internal deduplicated mixture of text and code
data, totaling 18T tokens balanced at a 1:1 text-to-code token ratio. The
code half spans over 100 programming languages, whose detailed composition is
reported in \autoref{tab:code-mix}.
Training and inference loop counts are matched throughout: a model
trained at $\loopcount = r$ is evaluated at $\loopcount = r$.
We use the Adam optimizer with $\beta_1 = 0.9$, $\beta_2 = 0.95$,
$\epsilon = 10^{-15}$, weight decay $0.1$, and gradient clipping at
norm $1.0$.
The learning rate is $\eta = 4 \times 10^{-4}$ with a cosine decay
schedule and a linear warmup over the first 5\% of training steps.
All runs use bf16 mixed precision with gradient checkpointing.
In total, training \modelname{} of different loops in this work consumed a total of 1M GPU hours.

\paragraph{Training Infrastructure.}
\label{sec:exp-infra}
We train PLT on a customized Megatron-LM stack with native support for
weight-tied loop unrolling.
The $\loopcount$ loops over the $L$-layer shared block are expanded into
$\loopcount\cdot L$ scheduled layers, but only the first loop instantiates
parameters; subsequent loops execute against references to the same modules, so
the parameter count, optimizer state, and checkpoint footprint remain those of a
single $L$-layer block regardless of $\loopcount$.
To keep weight sharing communication-free under pipeline parallelism, the
virtual-pipeline layout co-locates all $\loopcount$ instances of a given layer on
the same pipeline stage.
The first loop performs standard full attention and caches its keys and values;
each later loop issues two Transformer-Engine attention calls per layer, a
width-$w$ sliding-window attention over the current loop's keys and values and a
full attention over the frozen first-loop cache, combined by a per-head gate
(\autoref{eq:gswa}) that is zero-initialized to an even local/global blend.
Because the cross-loop offset reuses the first-loop cache and token embeddings,
these tensors are detached and their gradients accumulated through a custom
backward hook, preserving correct autograd under (virtual) pipeline scheduling.

\paragraph{Evaluation Protocol.}
\label{sec:exp-eval}

The four models are instruction-tuned with an identical supervised
fine-tuning recipe on 6M instruction-tuning examples and evaluated at their
matched training/inference loop
count, the baseline at $\loopcount = 1$ and the three PLT models at
$\loopcount = 2$, $\loopcount = 3$, and $\loopcount = 4$.
We assess each model on a broad external benchmark suite spanning code
generation, multilingual code, code reasoning, data-science and SQL, agentic
software engineering, and general tool use, using each benchmark's standard
protocol and metric.
We report the final supervised fine-tuning checkpoint for every model. Results,
together with comparisons to a range of open and proprietary models, are given
in \autoref{sec:exp-results}.

\subsection{Microscopic View: Per-Loop Diagnostic Lenses}
\label{sec:exp-micro}
The macroscopic view establishes that the loop count is decisive but is
silent on why: downstream accuracy is a single scalar that cannot explain
what happens inside the model.
To open up the computation, we dive into the model's internals and ask what
each loop contributes, and how the CLP offset modulates that
contribution.
Rather than relying on a single probe, we triangulate with three complementary
lenses, each interrogating a different stage of the forward pass.
A mechanism is credited only when the lenses agree.
Hidden-state dynamics examines the
representation \emph{as it is refined}, attention heat-map evolution examines
\emph{how information is routed}, and output-distribution shift examines the
\emph{prediction the refinement produces}.
Alongside the gain side captured by these lenses, we instrument the cost side
with an intrinsic offset cost that quantifies the CLP-induced positional
mismatch directly from the model's own states.


\subsubsection{Per-Loop Hidden-State Dynamics}
\label{sec:analysis-dynamics}

We track four statistics at each loop step $r$ to characterize the nature
and magnitude of the representational update.

\paragraph{Step size and angular change.}
The step size $\delta^{(r)} = \left\|\hidden^{(r)} - \hidden^{(r-1)}\right\|_2$ measures the magnitude of the hidden-state update at loop $r$, where
$\|\cdot\|_2$ denotes the Euclidean ($L_2$) norm used throughout.
The angular change is denoted as:
\begin{equation}
  \cos\theta^{(r)} = \frac{
    \left\langle \hidden^{(r)} - \hidden^{(r-1)},\;
                  \hidden^{(r-1)} - \hidden^{(r-2)} \right\rangle
  }{\left\|\hidden^{(r)} - \hidden^{(r-1)}\right\|_2
    \left\|\hidden^{(r-1)} - \hidden^{(r-2)}\right\|_2}
  \label{eq:angular}
\end{equation} where $\cos\theta^{(r)}$ is the update-direction alignment between two successive updates: $\cos\theta^{(r)}\!\approx\!1$ means consecutive loops keep refining in the same direction, $\cos\theta^{(r)}\!\approx\!0$ means orthogonal updates, and $\cos\theta^{(r)}\!<\!0$ signals direction reversal, i.e.\ oscillatory rather than convergent refinement \citep{pappone2025twoscale}.

\paragraph{Effective rank and fixed-point gap.}
The effective rank
\begin{equation}
  \mathrm{erank}(\hidden^{(r)}) =
  \exp\!\left(-\sum_i \bar\sigma_i \log \bar\sigma_i\right),
  \label{eq:erank}
\end{equation}
where $\bar\sigma_i$ are the normalized singular values of the $S \times d$
hidden-state matrix (computed on RMSNorm-normalized states so the measure is
scale-free), measures the geometric diversity of token representations at
loop $r$.
Representational diversity rises sharply from the embedding through the early
loops and \emph{peaks at loop~2}. A subsequent decline indicates that later
loops begin to narrow the representational subspace rather than enrich it,
eroding the model's capacity to maintain token-specific information
\citep{chen2026loop}.
The fixed-point gap directly measures how far the current state deviates from a fixed point of the shared block, providing a scalar summary of residual refinement capacity as below:
\begin{equation}
  \Delta_{\text{FP}}^{(r)} =
  \left\|\hidden^{(r)} - \block\!\left(\hidden^{(r)}\right)\right\|_2.
  \label{eq:fpgap}
\end{equation}

\paragraph{Intrinsic offset cost.}
Under the CLP mechanism, the input to loop $r \geq 2$ is
$B^{(r)} = \mathrm{Embed}(x) + \mathrm{shift}(\hidden^{(r-1)})$
(\autoref{eq:clp}), so token $i$ receives the loop-$(r{-}1)$ hidden state
of its neighbor $i{-}1$ rather than its own.
The degree to which this substitution distorts the input signal depends
directly on how dissimilar adjacent token representations are at that
loop boundary.
We therefore define the \emph{intrinsic offset cost} at loop $r$ as the mean
Euclidean distance between the representations of adjacent tokens at the
previous loop. This per-loop scalar $\Omega^{(r)}$ is computable directly from the neighboring hidden states of the LLM.
\begin{MiddleEquation}
\begin{equation}
  \Omega^{(r)} =
  \frac{1}{S}\sum_{i} \left\|\hidden^{(r-1)}_i - \hidden^{(r-1)}_{i-1}\right\|_2
  \label{eq:offset-cost}
\end{equation}
\end{MiddleEquation}where $S$ is the sequence length and a small $\Omega^{(r)}$
signals that representations have begun to homogenize, rendering the
shift nearly lossless.
Empirically $\Omega^{(r)}$ is \emph{nearly constant} across loops: adjacent
token representations remain comparably heterogeneous at every loop boundary,
so the CLP shift imposes a roughly fixed positional tax at each iteration.
Because the \emph{benefit} of an additional loop diminishes rapidly with depth
, this fixed cost
constitutes an ever-larger share of each loop's net effect, so that beyond a
small number of loops the offset penalty increasingly outweighs the shrinking
gain.
This interplay between a fixed offset cost and a diminishing loop gain is the
central mechanism examined in \autoref{sec:analysis-synthesis}.

\subsubsection{Attention Heat-Map Evolution Across Loops}
\label{sec:analysis-attn}

Attention patterns record how the model distributes information across token positions at each loop, and their evolution reveals whether successive loops specialize, engaging distinct subsets of token relationships, or degenerate into attentional redundancy.

\paragraph{Per-loop attention statistics.}
We track two scalar statistics per loop.
The attention entropy for head $h$ at query position $q$,
\begin{equation}
  \mathcal{H}^{(r,h)}_q = -\sum_{k=1}^S
    A^{(r,h)}_{qk} \log A^{(r,h)}_{qk},
  \label{eq:attn-entropy}
\end{equation}
where quantifies whether a head is globally diffuse or locally focused at loop $r$.
The inter-loop KL divergence
\begin{equation}
  D_{\text{KL}}^{(r)} = \frac{1}{HS}\sum_{h,q}
    \mathrm{KL}\!\left(A^{(r,h)}_q \;\|\; A^{(r-1,h)}_q\right)
  \label{eq:attn-div}
\end{equation}
which measures how much the attention distribution changes between consecutive loops.
A rapid decay of $D_{\text{KL}}^{(r)}$ toward zero indicates that
information routing has effectively frozen, and subsequent loops add no new
attention-level computation regardless of their hidden-state updates
\citep{lu2025latent}.

\paragraph{Attention-head diversity.}
To measure whether the attention heads remain specialized or collapse toward
redundant routing, we compute, at each loop, the \emph{effective rank} of the
$H$ per-head attention distributions at each query position, the entropy of
the normalized singular-value spectrum of the $H\times S$ matrix of head
attention vectors, together with the mean pairwise cosine similarity between
heads. A falling effective rank, equivalently a rising head similarity, signals
that the heads increasingly route information in the same way: the
attention-level analogue of the hidden-state effective-rank narrowing.

\paragraph{Local vs.\ global attention in G-SWA.}
In PLT, each non-first attention of loop is a gated mixture of a local
sliding-window component (over current-loop KV) and a global component
(over the frozen loop-1 KV cache) in \autoref{eq:gswa}.
We separately track the mean gate value $\bar{g}^{(r)}$ across heads and
positions. Under our convention (\autoref{eq:gswa}) $\bar{g}^{(r)}$ is the
weight placed on the \emph{global} branch (the frozen loop-1 KV cache), so
$\bar{g}^{(r)}\!\to\!1$ indicates near-total reliance on the loop-1 global
representation and $\bar{g}^{(r)}\!\to\!0$ indicates reliance on fresh local
context.
A gate that stays well above $0.5$ and changes little across loops indicates
that later loops keep drawing on the same frozen global cache rather than
constructing qualitatively new context.

\subsubsection{Output-Distribution Shift Across Loops}
\label{sec:analysis-output}

Applying the output head to intermediate hidden state $\hidden^{(r)}$ yields
a token-probability distribution
$p^{(r)} = \mathrm{Softmax}(\mathrm{Head}(\hidden^{(r)}))$.
Tracking $p^{(r)}$ across loops reveals whether the model progressively
refines a coarse initial prediction or whether its output distribution
oscillates or stagnates \citep{chen2026loop}.

\paragraph{Per-loop output-shift metrics.}
We employ three measures.
The Logit Lens rank \citep{lu2025latent}, the rank of the ground-truth
next token under $p^{(r)}$, indicates whether loop $r$ moves the model
closer to the correct prediction. A monotonically decreasing rank over $r$
constitutes the cleanest signature of iterative refinement.
The inter-loop KL divergence of output distributions,
\begin{equation}
  \Delta p^{(r)} = \mathrm{KL}\!\left(p^{(r)} \;\|\; p^{(r-1)}\right),
  \label{eq:output-shift}
\end{equation}
which measures the prediction change at each loop step.
The output entropy $\mathcal{H}(p^{(r)})$ tracks the confidence with which
the model commits to a prediction as loops accumulate.

\paragraph{Where refinement concentrates.}
Beyond aggregate trends, we ask how the post-context refinement ($r \geq 2$) is
divided among the refinement loops, measuring each loop's share under three
independent lenses: the magnitude of its output shift $\Delta p^{(r)}$, the
amount of attention re-routing it performs $D_{\mathrm{KL}}^{(r)}$, and the
fraction of tokens for which it is the peak-contribution loop
$r^* = \arg\max_r \Delta p^{(r)}$.
The first loop, which maps embeddings to contextual states, dominates the
unconditioned distribution and is excluded so as to isolate refinement.

\definecolor{tablegray}{gray}{0.92}
\begin{table}[t]
\centering
\caption{Comparison on code-generation and agentic / tool-use benchmarks.
Best per column in \textbf{bold}; our 7B models shaded. A single extra loop
($\loopcount{=}2$) is competitive with much larger systems, especially on
agentic tasks, while a second extra loop ($\loopcount{=}3$) regresses.
Avg. is the mean over available benchmarks (entries marked -- are excluded). HE+: HumanEval+;
MultiPL-E: multilingual avg.; BCB: BigCodeBench-Full;
LCB: LiveCodeBench; SWE: SWE-bench Verified; SWE-M: SWE-bench Multilingual;
TB-v1/v2: Terminal-Bench; M2W: Mind2Web; BFCL: tool use (v3).}
\label{tab:main-comparison}
\resizebox{\textwidth}{!}{%
\begin{tabular}{l ccccccccccc}
\toprule
\textbf{Model} & HE+ & MultiPL-E & BCB & LCB & SWE & SWE-M & TB-v1 & TB-v2 & M2W & BFCL & Avg. \\
\midrule
\multicolumn{12}{l}{\textit{Small Open models, $\leq$\,32B}} \\
DeepSeek-Coder-V2-Lite-Instruct & 75.6 & 71.5 & 37.8 & 19.4 & 0.0 & 0.0 & 5.0 & 0.0 & 26.7 & -- & 26.2 \\
Qwen2.5-Coder-7B-Instruct       & 81.7 & 75.4 & 37.8 & 18.9 & 0.0 & 0.0 & 6.3 & 0.0 & 38.4 & 54.2 & 31.3 \\
Seed-Coder-8B-Instruct          & 75.6 & 75.1 & 44.6 & 22.3 & 0.0 & 0.0 & 7.5 & 2.5 & 38.2 & -- & 29.5 \\
Qwen2.5-Coder-14B-Instruct      & 59.8 & 78.8 & 47.0 & 24.6 & 0.0 & 0.0 & 8.8 & 0.0 & 42.7 & 59.9 & 32.2 \\
Qwen2.5-Coder-32B-Instruct      & 86.6 & 79.6 & 48.0 & 27.4 & 0.0 & 0.0 & 5.0 & 4.5 & 32.5 & 62.3 & 34.6 \\
\midrule
\multicolumn{12}{l}{\textit{Large open models}} \\
Qwen3-235B-A22B-Instruct-2507   & 91.5 & 87.9 & 47.4 & 51.8 & 45.2 & -- & 15.0 & 13.5 & 49.0 & 71.2 & 52.5 \\
Kimi-Dev-72B                    & 86.0 & 80.3 & 45.4 & 40.0 & 60.4 & -- & -- & 2.3 & -- & 55.5 & 52.8 \\
Kimi-K2-Instruct-0905           & 89.6 & 85.7 & 49.8 & 53.7 & 69.2 & 33.5 & 44.5 & 27.8 & 53.4 & 70.3 & 57.8 \\
Qwen3-Coder-480B-A35B-Instruct  & 92.7 & 87.4 & 49.4 & 53.9 & 67.0 & 32.7 & 37.5 & 23.6 & 54.0 & 68.7 & 56.7 \\
DeepSeek-V3.2                   & 88.4 & 85.8 & 48.1 & 83.3 & 73.1 & 37.4 & 23.8 & 46.4 & 47.2 & 68.8 & 60.2 \\
GLM-4.7                         & 79.9 & 69.0 & 45.7 & 84.9 & 73.8 & -- & 36.3 & 41.0 & 53.7 & 64.8 & 61.0 \\
\midrule
\multicolumn{12}{l}{\textit{Proprietary models}} \\
GPT-5.1                         & 90.0 & 86.2 & 46.8 & 87.0 & 76.3 & -- & 35.0 & 47.6 & 55.1 & 64.4 & 65.4 \\
Claude-Opus-4.5                 & 93.3 & 91.0 & \textbf{53.3} & 87.1 & \textbf{80.9} & \textbf{50.0} & \textbf{47.5} & \textbf{59.3} & 57.9 & \textbf{78.9} & \textbf{69.9} \\
Gemini-3-Pro                    & \textbf{94.5} & \textbf{91.2} & 47.1 & \textbf{91.7} & 76.2 & 42.7 & 46.3 & 54.2 & \textbf{60.3} & 78.2 & 68.2 \\
\midrule
\multicolumn{12}{l}{\textit{Ours (7B)}} \\
\rowcolor{tablegray} Baseline ($\loopcount{=}1$)        & 81.1 & 69.5 & 40.1 & 27.4 & 43.0 & 14.0 & 26.3 & 11.2 & 35.3 & 32.2 & 38.0 \\
\rowcolor{tablegray} \modelname{} ($\loopcount{=}2$)    & \textbf{84.1} & \textbf{73.9} & \textbf{46.1} & \textbf{35.4} & \textbf{64.4} & \textbf{31.0} & \textbf{34.2} & \textbf{21.0} & \textbf{34.5} & \textbf{40.1} & \textbf{46.5} \\
\rowcolor{tablegray} \modelname{} ($\loopcount{=}3$)    & 75.0 & 69.8 & 43.3 & 28.6 & 27.6 & 11.0 & 30.0 & 12.2 & 35.1 & 36.3 & 36.9 \\
\rowcolor{tablegray} \modelname{} ($\loopcount{=}4$)    & 76.8 & 67.3 & 40.8 & 24.5 & 22.4 & 9.3 & 26.3 & 9.0 & 41.4 & 39.5 & 34.3 \\
\bottomrule
\end{tabular}}
\end{table}

\section{Per-Loop Interpretability Analysis}
\label{sec:analysis}

\subsection{Main Results}
\label{sec:exp-results}

\autoref{tab:main-comparison} compares our models with a range of open and
proprietary systems on a representative benchmark subset.
Two observations stand out.
First, performance is strongly \emph{non-monotonic} in the loop count: a single
additional loop ($\loopcount = 2$) improves markedly over the non-looped
baseline, whereas a second extra loop ($\loopcount = 3$) regresses, often below
the baseline.
Second, this single extra loop makes our 7B model strikingly competitive with
far larger systems, most notably on SWE-bench Verified, where it reaches
$64.4\%$, surpassing 30B--72B open models and approaching 480B-scale ones.
The same configuration also attains $33.4\%$ on the agentic SWE-bench-CC,
confirming that the loop-2 gains carry over to held-out agentic settings.
The non-monotonic curve, peaking after one additional loop, is the phenomenon
whose representational origin the rest of the paper investigates.

\begin{figure}[h]
\centering
\includegraphics[width=\linewidth]{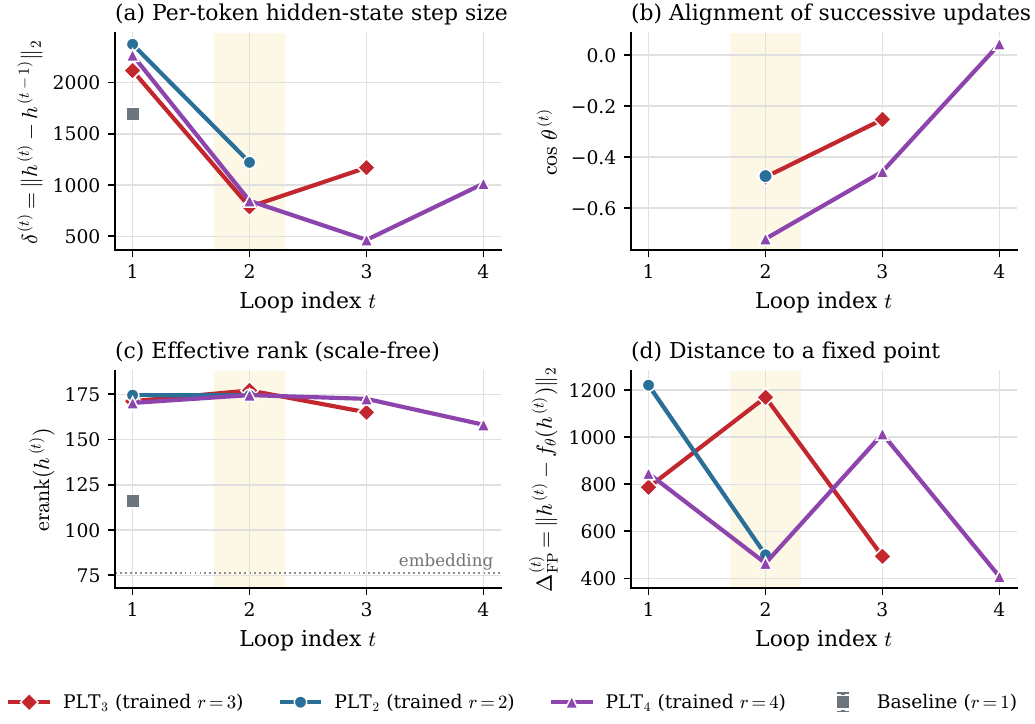}
\caption{Step size $\delta^{(r)}$ (top-left), angular change $\cos\theta^{(r)}$
(top-right), scale-free effective rank $\mathrm{erank}(\hidden^{(r)})$
(bottom-left), and fixed-point gap $\Delta_{\text{FP}}^{(r)}$ (bottom-right)
as a function of loop index $r$. Lines: PLT$_2$/PLT$_3$/PLT$_4$
(trained $\loopcount{=}2,3,4$); the baseline ($\loopcount{=}1$) is shown
where defined. Shaded bands are 95\% CIs over 500 samples (often narrower than
the markers); the dotted line in (c) marks the embedding. Effective rank peaks
at loop~2 and declines for every deeper loop; successive updates are oscillatory
($\cos\theta^{(r)}<0$) through the refinement loops, and the step size shrinks to
a mid-depth minimum before rebounding at the final (output) loop.}
\label{fig:dynamics}
\end{figure}

\begin{figure}[h]
\centering
\includegraphics[width=0.66\linewidth]{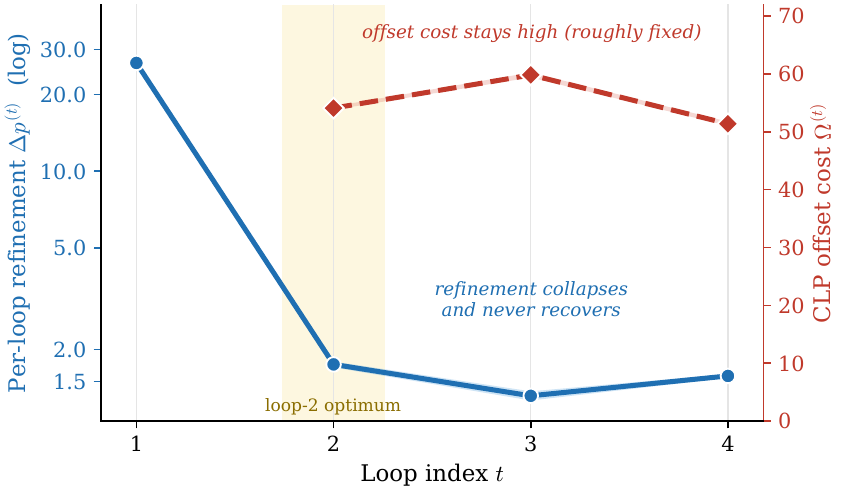}
\caption{The gain--cost scissors (PLT$_4$). The per-loop refinement gain
$\Delta p^{(r)}$ (output-distribution KL; left axis, log) collapses after loop~2
and never recovers, whereas the intrinsic CLP offset cost $\Omega^{(r)}$
(\autoref{eq:offset-cost}; right axis) stays high and roughly fixed. At every
extra loop the offset cost exceeds the per-loop gain by $30$--$45\times$, so the
fixed offset tax dominates the shrinking refinement beyond loop~2. 500 samples;
shaded bands are 95\% CIs.}
\label{fig:offset-cost}
\end{figure}

\begin{figure}[t]
\centering
\includegraphics[width=\linewidth]{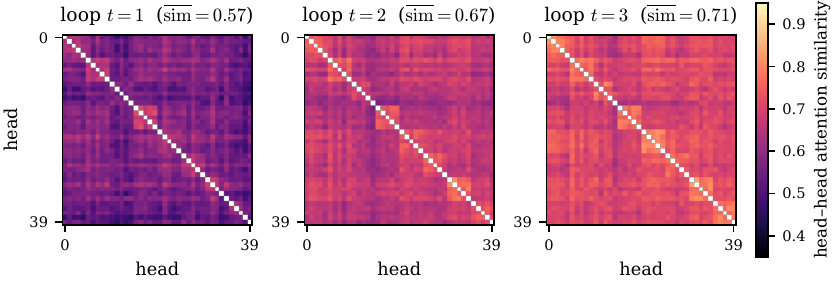}
\caption{Head$\times$head cosine similarity of the per-head attention
distributions at loops $r=1,2,3$ (PLT$_3$, 500 held-out samples; self-similarity
on the diagonal masked). Brighter cells indicate more redundant heads;
$\overline{\mathrm{sim}}$ is the mean off-diagonal similarity. Heads grow
progressively more redundant across loops.}
\label{fig:attn-heatmaps}
\end{figure}

\begin{figure}[t]
\centering
\includegraphics[width=\linewidth]{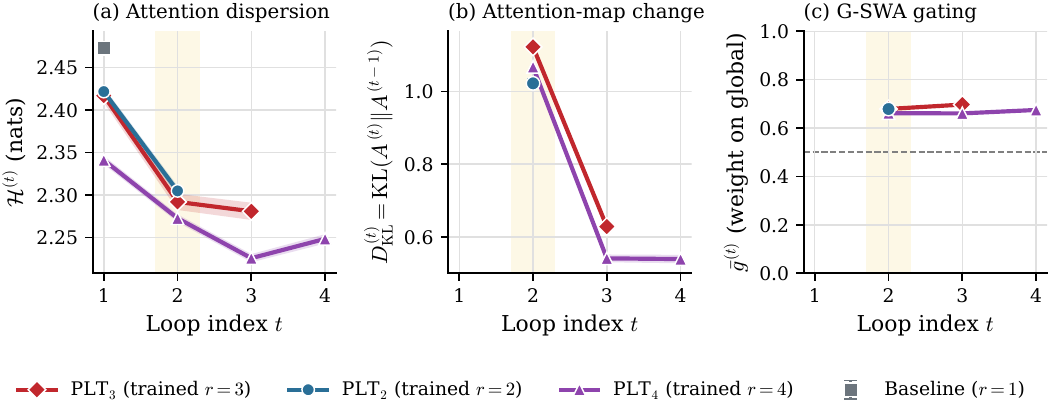}
\caption{Mean attention entropy $\mathcal{H}^{(r)}$ (left),
inter-loop KL divergence $D_{\text{KL}}^{(r)}$ (middle),
and mean G-SWA gate $\bar{g}^{(r)}$ (right; the weight on the global loop-1
branch) as a function of loop index $r$ (PLT$_2$/PLT$_3$/PLT$_4$). The inter-loop
KL drops sharply after loop~2 and stays low, indicating that attention routing
largely freezes once the second loop completes; the gate stays well above
$0.5$ at every loop.}
\label{fig:attn-stats}
\end{figure}

\begin{figure}[t]
\centering
\includegraphics[width=\linewidth]{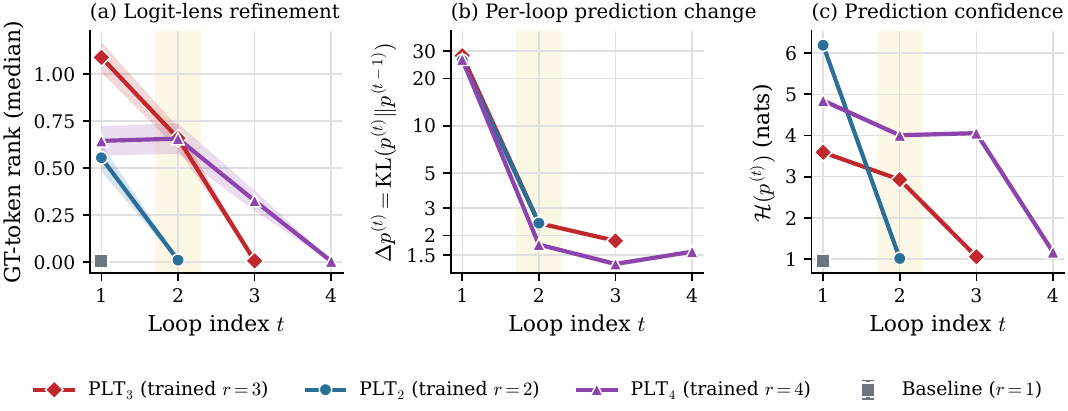}
\caption{Logit-lens ground-truth token rank (left), inter-loop output
KL divergence $\Delta p^{(r)}$ (middle, log scale), and output entropy
$\mathcal{H}(p^{(r)})$ (right) as a function of loop index $r$
(PLT$_2$/PLT$_3$/PLT$_4$).
Predictions sharpen monotonically with depth, but the \emph{per-loop} change
$\Delta p^{(r)}$ collapses after loop~2 (the small uptick at the final loop is
output readout rather than new refinement), marking the onset of diminishing
returns.}
\label{fig:output-shift}
\end{figure}

\begin{figure}[t]
\centering
\includegraphics[width=0.92\linewidth]{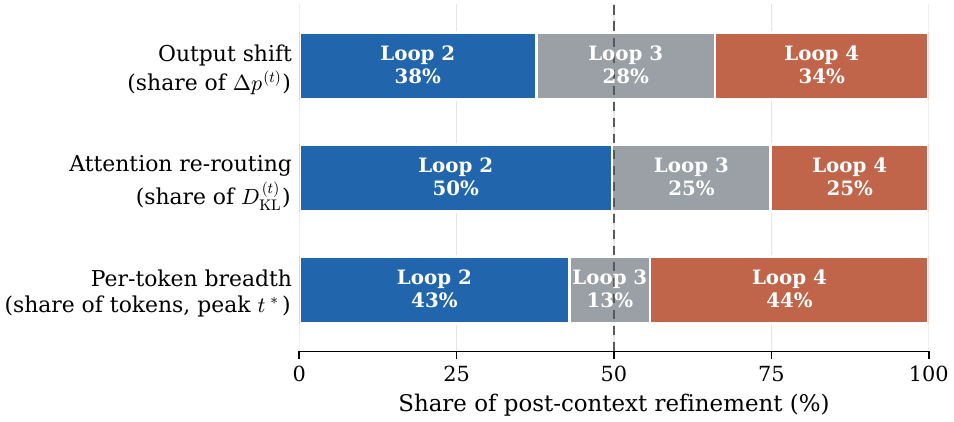}
\caption{How post-context refinement is distributed across the extra loops of
PLT$_4$. Each bar splits the refinement carried by the refinement loops
($r \geq 2$) into its loop-2, loop-3, and loop-4 shares, under three independent
lenses: the output shift $\Delta p^{(r)}$, the inter-loop attention re-routing
$D_{\mathrm{KL}}^{(r)}$, and the per-token peak-contribution loop
$r^* = \arg\max_r \Delta p^{(r)}$. The middle extra loop (loop~3, grey) carries
the \emph{smallest} share on every lens: genuine refinement concentrates at
loop~2,where attention re-routing and effective rank also peak,while loop~4's
large output-side share is final readout rather than representational
enrichment (its effective rank is the lowest; \autoref{fig:dynamics}c).
(Loop~1 builds context and is excluded to isolate refinement.)}
\label{fig:peak-loop}
\end{figure}

\subsection{Synthesis: Loop Contribution vs.\ Offset Cost}
\label{sec:analysis-synthesis}

We report per-loop hidden dynamics as mentioned in \autoref{sec:analysis-dynamics} with \autoref{fig:dynamics}  and \autoref{fig:offset-cost}. We then visualize attention-head evolution as mentioned in \autoref{sec:analysis-attn} with
\autoref{fig:attn-heatmaps} and \autoref{fig:attn-stats}. Finally, we show output distribution shift as mentioned in \autoref{sec:analysis-output} with \autoref{fig:output-shift} and \autoref{fig:peak-loop}.

\paragraph{Loop 2 is the principal site of productive refinement.}
The first loop performs the largest transformation, mapping embeddings to
contextual states, but among the \emph{refinement} loops the second loop
carries the most meaningful change: it produces the largest inter-loop attention
divergence $D_{\text{KL}}^{(2)}$ and the largest per-loop output shift
$\Delta p^{(2)}$, and it is where effective rank peaks.
Representational diversity is thus maximized at loop~2, and every deeper
refinement loop only narrows it.

\paragraph{Beyond loop 2: diminishing and non-productive returns.}
Past the second loop the marginal contribution of each iteration collapses:
$\Delta p^{(r)}$ and $D_{\text{KL}}^{(r)}$ drop sharply, and the effective rank
\emph{declines} from its loop-2 peak, signalling that further loops narrow the
representational subspace rather than enrich it.
The attention heads echo this at the routing level: their diversity falls and
they grow increasingly redundant from loop to loop
(\autoref{fig:attn-heatmaps}).
The hidden-state updates beyond loop~2 are either near-inert or
\emph{oscillatory}: successive update directions reverse ($\cos\theta^{(r)}<0$),
so the extra computation reflects non-convergent movement or final readout
rather than genuine refinement.
In the four-loop model this is starkest at the middle extra loop, which acts as
a near-dead pass-through, while the final loop merely re-reads the prediction
(\autoref{fig:peak-loop}).

\paragraph{The CLP offset is a fixed per-loop tax.}
The intrinsic offset cost $\Omega^{(r)}$ (\autoref{eq:offset-cost}) is
approximately constant across loops: the CLP shift injects a comparable
positional mismatch at every loop boundary.
Because the benefit of each additional loop diminishes rapidly, this fixed cost
claims a growing share of the net effect, so that beyond the second loop the
offset penalty increasingly dominates the shrinking gain.
Together with the post-peak narrowing of effective rank, this offers a
mechanistic account of why performance peaks at $\loopcount=2$ and degrades
with further loops.

\begin{table}[h]
\centering
\caption{Per-loop behavioral signatures in the four-loop model (PLT$_4$),
averaged over 500 held-out samples: per-token step size $\delta^{(r)}$,
output-distribution shift $\Delta p^{(r)}$, effective rank, and update
alignment $\cos\theta^{(r)}$ at each refinement loop. \underline{Underline}
marks the largest value in each column.}
\label{tab:behavior-summary}
\begin{tabular}{lcccc}
\toprule
\textbf{Loop} & $\delta^{(r)}$ & $\Delta p^{(r)}$ & \textbf{Eff.\ rank} & $\cos\theta^{(r)}$ \\
\midrule
$r=2$ & $\underline{846}$            & $\underline{1.75}$  & $\underline{174.6}$ & $\underline{-0.72}$ \\
$r=3$ & $464$            & $1.32$              & $172.5$             & $-0.46$ \\
$r=4$ & $1014$ & $1.58$            & $158.2$             & $0.04$ \\
\bottomrule
\end{tabular}
\end{table}

\subsection{Explicit and Latent Chains of Thought are Complementary}
\label{sec:analysis-cot}

The looped computation analyzed above can be read as a form of \emph{latent}
chain-of-thought: the model performs iterative refinement in representation
space across loops without emitting any intermediate tokens
\citep{geiping2025scaling,lu2025latent}.
A reasoning (``thinking'') model, by contrast, externalizes an \emph{explicit}
chain-of-thought as output tokens.
Because a single additional loop ($\loopcount = 2$) is the optimal operating
point (\autoref{sec:exp-results}), we ask whether these two reasoning
channels, namely explicit token-level CoT and latent loop-level refinement,
are redundant or complementary.
We compare, at this same $\loopcount = 2$ configuration, the instruction-tuned
model (latent loop only) against its thinking counterpart, a variant fine-tuned
to emit an explicit reasoning trace (explicit CoT atop the latent loop), on
reasoning-intensive benchmarks.

\begin{table}[h]
\centering
\caption{Instruction-tuned vs.\ thinking model at the optimal single extra loop
($\loopcount = 2$).}
\label{tab:cot-synergy}
\begin{tabular}{lccccc}
\toprule
\textbf{Model ($\loopcount = 2$)} & LCB & CRUX & MultiPL-E & FullStackBench & BCB-Hard \\
\midrule
Instruct (latent loop only)        & 35.4 & 86.9 & 73.9 & 47.2 & 23.7 \\
Thinking (explicit CoT $+$ loop)   & \textbf{62.3} & \textbf{93.5} & \textbf{77.8} & \textbf{49.9} & \textbf{26.4} \\
\midrule
$\Delta$                           & $+26.9$ & $+6.6$ & $+3.9$ & $+2.7$ & $+2.7$ \\
\bottomrule
\end{tabular}
\end{table}

\autoref{tab:cot-synergy} shows that on reasoning-heavy tasks the thinking
variant improves over the instruction-tuned variant by a wide margin, most
strikingly on LiveCodeBench ($+26.9$ points), far exceeding the gain
attributable to either ingredient in isolation: explicit CoT alone does not
improve the non-looped model on these tasks, and the loop alone yields only
single-digit gains for the instruction-tuned model.
The combination is therefore \emph{super-additive}: the joint gain exceeds the
sum of the gains contributed by explicit reasoning and latent recurrence
separately.
We attribute this to the two mechanisms operating at different granularities:
the explicit CoT decomposes a problem into intermediate textual steps, while the
latent loop refines the representation that underlies each step, so that
chaining looped refinements through an explicit reasoning trace compounds their
individual contributions.
Latent depth recurrence and explicit reasoning thus appear to be complementary
axes of test-time computation rather than substitutes, and their interaction is
strongest precisely at the single-extra-loop operating point identified by our
per-loop analysis.

\section{Discussion}
\label{sec:optimize_loop_strategy}

The per-loop analysis in \autoref{sec:analysis} reveals a consistent
pattern: loop contributions are not uniform, and the interaction between
representational gains and the cost imposed by the CLP offset shifts across
loop index in a way that explains the non-monotonic performance curve.

\paragraph{Loop 2 is the principal site of productive refinement.}
The second loop is the primary site of productive refinement: among the
refinement loops it introduces the most coherent hidden-state update, the
highest inter-loop attention divergence, and the greatest output-distribution
shift, and it is where effective rank peaks.
The first loop establishes the global KV cache used by all subsequent loops,
so the quality of the loop-1 representations sets a ceiling on the
information available to every later loop via the frozen global attention
branch.
The second loop then refines these representations using both fresh local
context and the global cache, producing the largest net improvement.

\paragraph{Beyond loop 2: diminishing gains against a fixed cost.}
Beyond loop 2, the model faces compounding constraints.
The effective rank declines from its loop-2 peak (representations become less
diverse), so the shared block operates on a progressively lower-dimensional
input, reducing the capacity for new computation. Meanwhile, the hidden-state
update becomes oscillatory rather than convergent.
At the same time, the intrinsic offset cost $\Omega^{(r)}$ remains roughly
constant across loops: the CLP shift injects a comparable positional mismatch
at every boundary.
Because per-loop gains shrink rapidly while this offset cost stays fixed, the
mismatch claims an ever-larger share of each loop's net effect, so that beyond
the second loop the cost increasingly outweighs the benefit.

\paragraph{Practical guidelines for loop-count selection.}
These findings suggest several practical guidelines.
The performance-saturation point identified in the analysis corresponds to
a natural operating threshold for PLT deployment: $\loopcount = 2$ captures the
dominant refinement step while incurring only a single additional forward pass.
For finer-grained loop-count selection, the effective-rank trajectory is a
lightweight per-model diagnostic that requires no exhaustive sweep: if effective
rank is still rising at the candidate loop (representational diversity is not
yet saturated), an additional loop may yield genuine refinement, whereas a rank
that has begun to fall signals the onset of narrowing, after which further loops
mostly add the fixed CLP offset cost without compensating gain.

\section{Related Work}
\label{sec:related}

\subsection{Foundations of Looped Transformers}
\label{sec:related-foundations}

The Universal Transformer (UT) \citep{dehghani2018universal} serves as the
canonical starting point for looped LLM research.
UT shares a single transformer block across depth and augments it with
per-position adaptive computation time (ACT), a dynamic halting mechanism
that allows individual tokens to exit the loop at different depths.
Under certain conditions UT is provably Turing-complete, a theoretical
property not generally attributable to fixed-depth transformers.
The programmable-computer interpretation of looped transformers was made
precise by \citet{giannou2023looped}, who demonstrated that a constant number
of encoder layers in a loop suffices to emulate a general-purpose
instruction-set computer, including in-context learning via
back-propagation.
\citet{yang2023looped} further established that looped transformers match
standard transformers on in-context learning benchmarks while using fewer than
10\% of the parameters, affirming their practical viability.

\subsection{Test-Time Compute Scaling via Depth Recurrence}
\label{sec:related-scaling}

A central motivation for recent work on looped LLMs is \emph{test-time compute
scaling}: rather than generating additional tokens (as in chain-of-thought
reasoning), a model can be run for more loops on harder inputs, performing
implicit multi-step reasoning in latent space.
\citet{geiping2025scaling} demonstrated this at scale with Huginn-3.5B, a
depth-recurrent transformer pretrained on 800B tokens that improves on
reasoning benchmarks by applying up to 50 loops at inference, achieving an
effective compute budget equivalent to a 50B-parameter model.
Notably, this approach requires no specialized training data and can operate
within smaller context windows than chain-of-thought methods. This latent reasoning paradigm, however, imposes a direct cost:
each additional loop constitutes a full forward pass through the shared
block, multiplying sequential inference latency by loopcount.
More critically, standard KV-cache implementations store keys and values
per layer \emph{per loop}, causing memory to grow as $O(loopcount \cdot L)$
for a block of $L$ layers, a factor of loopcount overhead relative to the
single-pass footprint.
With too many loopcount, this overhead renders many architectures
impractical for on-device or memory-constrained deployment.
\citep{vendrell2026memoryefficient}.

\subsection{Memory and Latency Reduction Techniques}
\label{sec:related-efficiency}

Several works aim to retain the benefits of deep looping while reducing its inference cost. MELT decouples recurrence depth from memory by maintaining a single shared KV cache per layer across loops, updated through a learnable gating mechanism \citep{vendrell2026memoryefficient}. PLT instead targets latency by breaking sequential inter-loop dependencies with Cross-Loop Parallelism (CLP), and combines this with Gated Sliding-Window Attention (G-SWA) over shared global KV states and local current-loop context to keep memory nearly constant \citep{wu2025parallel}. LT2 further reduces the cost of looped inference by replacing quadratic softmax attention with linear or sparse attention variants, leveraging recurrence for iterative memory refinement \citep{deng2026lt2}. These methods improve the efficiency of looped computation, while our work focuses on the complementary question of how such efficiency mechanisms, especially PLT's CLP offset, affect the usefulness of additional loops.

\subsection{Architectural Variants}
\label{sec:related-variants}

Beyond efficiency-oriented designs, another line of work explores richer looped architectures. Some methods relax strict weight sharing by adding lightweight loop-specific adaptation, such as depth-wise LoRA adapters for converting pretrained LLMs into recursive transformers \citep{bae2024relaxed}. Others restructure the recurrent computation itself, including Hyperloop Transformers with begin-middle-end partitioning and hyper-connections for inter-loop mixing \citep{zeitoun2026hyperloop}, CART with a context-anchored recurrent core that cross-attends to frozen precomputed context tensors \citep{capps2026cart}, and HRM-LM with fast and slow modules operating at different loop timescales \citep{han2026hierarchical}. Additional variants increase capacity or stability through mixture-of-experts feedforward layers \citep{csords2024moeut}, fixed-point refinement with attractor modules \citep{feinashley2026solve}, or post-training conversion of standard LLMs into looped encoder-reasoner-decoder architectures \citep{park2026loopus}. These works broaden the design space of looped Transformers, whereas our focus is complementary: we study how loop count affects the behavior of an efficient PLT architecture and why its performance saturates.

\subsection{Scaling Laws and Representation Dynamics}
\label{sec:related-dynamics}

Recent work has begun to study how looped models scale and how their internal representations evolve with recurrence depth. Scaling-law analyses show that increasing loop count yields diminishing returns: \citet{schwethelm2026how} estimate that looping a block $r$ times is worth only $r^{0.46}$ unique parameters in validation loss, far below full equivalence to adding new layers. Complementary studies examine recurrent representation dynamics, finding that loop updates can become smaller, more orthogonal, or structured across multiple timescales \citep{pappone2025twoscale}. Other work highlights stability as a central limitation: \citet{yang2026stabilizing} show that performance can peak at an intermediate loop depth and then collapse, and propose fixed-point regularization to stabilize recurrent computation. On the interpretability side, prior analyses probe whether deeper recurrence corresponds to meaningful latent reasoning or natural-language-like intermediate computation, often finding mixed evidence and signs of representational degradation \citep{chen2026loop,lu2025latent}. Our work is closest in spirit to these representation-dynamics studies, but focuses specifically on PLT: we analyze how its efficiency mechanism changes the gain--cost profile across loops and why saturation occurs at a low loop count.

\section{Conclusion}
\label{sec:conclusion}

Looped Transformers provide an appealing mechanism for scaling latent computation without increasing parameter count, but their behavior under increasing loop count remains poorly understood. In this work, we study this problem in PLT through a gain--cost perspective: each additional loop may provide useful refinement, but the CLP offset also introduces a structural positional mismatch at every loop boundary. Our controlled loop-wise analysis shows that the second loop is the primary source of productive refinement, producing meaningful changes in hidden states, attention routing, and output distributions, while later loops yield diminishing and increasingly oscillatory updates. Because the CLP-induced mismatch remains approximately constant as marginal loop gains shrink, additional loops eventually become unproductive, explaining why PLT saturates at a small loop count. These findings provide interpretability-grounded diagnostics for loop-count selection without exhaustive benchmark sweeps, and suggest future directions such as adaptive offset mechanisms, dynamic loop allocation, and a deeper understanding of how latent recurrence interacts with explicit chain-of-thought reasoning.

\clearpage

\bibliography{ref}

\appendix

\section{Forward-Pass Pseudocode}
\label{app:pseudocode}

\autoref{alg:plt-unified} gives the PLT forward pass ($w = 64$
throughout).

\begin{algorithm}[H]
\caption{PLT Forward Pass ($w=64$)}
\label{alg:plt-unified}
\begin{algorithmic}[1]
\Input tokens $x$, loop count $\loopcount$
\Output logits $y$
\State $\hidden^{(0)} \leftarrow \mathrm{Embed}(x)$
\For{$r = 1$ \textbf{to} $\loopcount$}
  \If{$r = 1$}
    \State $\hidden^{(1)} \leftarrow \block\!\left(\hidden^{(0)}\right)$
      \Comment{standard full-context attention}
    \State $K_{\text{share}}, V_{\text{share}} \leftarrow \mathrm{KV}\!\left(\hidden^{(1)}\right)$
      \Comment{loop-1 KV cache, shared with all subsequent loops}
  \Else
    \State $B^{(r)} \leftarrow \mathrm{Embed}(x) + \mathrm{shift}\!\left(\hidden^{(r-1)}\right)$
      \Comment{CLP: right-shift by 1 position}
    \State $y_{\text{global}} \leftarrow \mathrm{Attn}\!\left(Q(B^{(r)}),\, K_{\text{share}},\, V_{\text{share}}\right)$
    \State $y_{\text{local}} \leftarrow \mathrm{SWAttn}\!\left(Q(B^{(r)}),\, K(B^{(r)}),\, V(B^{(r)}),\, 64\right)$
    \State $g^{(r)} =
\sigma\!\left(f_{\mathrm{gate}}(\mathrm{RMSNorm}(B^{(r)}))\right).$
      \Comment{G-SWA fusion (\autoref{eq:gswa})}
    \State $\hidden^{(r)} \leftarrow \mathrm{FFN}\!\left(\tilde{y}\right)$
  \EndIf
\EndFor
\State $y \leftarrow \mathrm{Head}\!\left(\hidden^{(\loopcount)}\right)$
\end{algorithmic}
\end{algorithm}

\section{Model Architecture Configurations}
\label{app:model-config}

\begin{table}[H]
\centering
\caption{Base model configuration used across all experiments.}
\label{tab:model-config}
\begin{tabular}{ll}
\toprule
\textbf{Hyperparameter} & \textbf{Value} \\
\midrule
Layers $L$                        & 14 \\
Hidden size $d$                   & 5120 \\
Attention heads $H$               & 40 \\
KV groups (GQA)                   & 8 \\
Head dimension                    & 128 \\
FFN intermediate size             & 27{,}648 \\
Activation                        & SwiGLU \\
Normalization                     & RMSNorm ($\epsilon = 10^{-5}$) \\
Position embedding                & RoPE (base $5\times10^{5}$) \\
Attention                         & Flash Attention, no dropout \\
QK LayerNorm                      & no \\
Bias in linear layers             & no \\
Precision                         & bf16 \\
Vocabulary size                   & 76{,}800 \\
Total parameters                  & $\approx$7B \\
Training tokens                   & 18T \\
Window size $w$                   & 64 (fixed) \\
CLP offset                        & enabled \\
Loop counts $\loopcount$ analyzed & 1, 2, 3, 4 \\
\bottomrule
\end{tabular}
\end{table}

\noindent
The per-loop interpretability analysis (\autoref{sec:analysis}) is
conducted across all loop counts, $\loopcount \in \{1, 2, 3, 4\}$.


\section{Pretraining Code-Data Composition}
\label{app:code-mix}

The pretraining corpus is balanced at a 1:1 text-to-code token ratio
(\autoref{sec:exp-train}). \autoref{tab:code-mix} breaks down the code half by
programming language: the ten largest languages by token share, with the
remaining 93 languages aggregated into ``Others''. Shares are computed over
code tokens only.

\begin{table}[h]
\centering
\caption{Composition of the code portion of the 18T-token pretraining mixture,
as token share of all code tokens. Top-10 languages shown individually; the
remaining 93 languages are grouped.}
\label{tab:code-mix}
\begin{tabular}{lr}
\toprule
\textbf{Language} & \textbf{Token share (\%)} \\
\midrule
Java                & 10.3 \\
Python              & 10.1 \\
JavaScript          & 9.4 \\
Markdown            & 8.7 \\
TypeScript          & 8.3 \\
C                   & 5.2 \\
C\texttt{++}        & 5.0 \\
PHP                 & 4.7 \\
C\#                 & 4.0 \\
HTML                & 3.7 \\
\midrule
Others (93 languages) & 30.5 \\
\bottomrule
\end{tabular}
\end{table}

\end{document}